\documentclass{optica-article}

\journal{opticajournal} % for journals or Optica Open

\articletype{Research Article}

\usepackage{lineno}
\usepackage[colorinlistoftodos]{todonotes}

\begin{document}

\title{Pose-Free 3D Quantitative Phase Imaging of Flowing Cellular Populations} 

\author{Enze Ye,\authormark{2,3,\dag} Wei Lin,\authormark{1,\dag} Shaochi Ren,\authormark{2,3} Yakun Liu,\authormark{1} Xiaoping Li, \authormark{4} Hao Wang, \authormark{5,6} He Sun,\authormark{2,3,*} and Feng Pan\authormark{1,*} }

\address{
\authormark{1}School of Instrumentation Science \(\&\) Optoelectronics Engineering, Beihang University, Beijing, \\
\authormark{2}College of Future Technology, Peking University, Beijing, China\\
\authormark{3}National Biomedical Imaging Center, Peking University, Beijing, China\\
\authormark{4}Department of Obstetrics and Gynecology, Peking University People’s Hospital, Beijing, China\\
\authormark{5}Peking University Third Hospital, Department of Radiation Oncology, Beijing, China.\\
\authormark{6}Peking University Third Hospital, Cancer Center, Beijing, China\\
\authormark{\dag}The authors contributed equally to this work.\\
\authormark{*}The corresponding authors (hesun@pku.edu.cn, panfeng@buaa.edu.cn).
}
%% email address is required; see note below about the corresponding author designation

% use {asbstract*} to suppress the copyright line. Copyright information will be added in production

\begin{abstract*} 

High-throughput 3D quantitative phase imaging (QPI) in flow cytometry enables label-free, volumetric characterization of individual cells by reconstructing their refractive index (RI) distributions from multiple viewing angles during flow through microfluidic channels. However, current imaging methods assume that cells undergo uniform, single-axis rotation, which require their poses to be known at each frame. This assumption restricts applicability to near-spherical cells and prevents accurate imaging of irregularly shaped cells with complex rotations. As a result, only a subset of the cellular population can be analyzed, limiting the ability of flow-based assays to perform robust statistical analysis. We introduce OmniFHT, a pose-free 3D RI reconstruction framework that leverages the Fourier diffraction theorem and implicit neural representations (INRs) for high-throughput flow cytometry tomographic imaging. By jointly optimizing each cell’s unknown rotational trajectory and volumetric structure under weak scattering assumptions, OmniFHT supports arbitrary cell geometries and multi-axis rotations. Its continuous representation also allows accurate reconstruction from sparsely sampled projections and restricted angular coverage, producing high-fidelity results with as few as 10 views or only 120° of angular range. OmniFHT enables, for the first time, in situ, high-throughput tomographic imaging of entire flowing cell populations, providing a scalable and unbiased solution for label-free morphometric analysis in flow cytometry platforms.

\end{abstract*}

\section{Introduction}

Quantitative phase imaging (QPI) is a label-free optical technique that enables high-contrast visualization of transparent biological cells by measuring light phase shifts induced by refractive index (RI) variations~\cite{Zangle2014}. Unlike brightfield microscopy, whose contrast is limited by minimal absorption in unlabeled cells, and fluorescent microscopy, which induces phototoxicity and photobleaching, QPI provides quantitative imaging without the need for staining or labeling—enabling long-term observation of live cells in their native states~\cite{Park2018,Colomb2006,Ferraro2011,Cotte2013,Kemper2010,kim2018label}. 

Three-dimensional (3D) QPI techniques, such as holographic tomography (HT), further combine multi-angle complex-field measurements with tomographic inversion to reconstruct the sample’s volumetric RI distribution, enabling label-free, high-contrast 3D morphological and biophysical analysis~\cite{Charriere:06, MEROLA2018, Kim2024}. However, current 3D QPI methods typically rely on controlled sample rotation or illumination scanning, restricting them to cell cultures in static imaging dishes, precluding their use in dynamic flow for high-throughput analysis. To address this limitation, recent work on in-flow holographic tomography (FHT) integrates HT with microfluidic flow cytometry~\cite{Sung2014, Merola2017, PIRONE2021}. As cells flow through microchannels, hydrodynamic shear rotates them, naturally providing multiple viewing angles without mechanical tip-tilt or external beam steering. This enables continuous, high-throughput acquisition of projection images from thousands of cells per minute, facilitating rapid, label-free phenotyping with direct applications in clinical diagnostics, drug screening, and rare cell detection~\cite{Pirone2023, Funamizu2018}.

Despite recent advances, one central challenge in FHT is accurately determining of the rotation angle for each projection image, since pose errors introduce reconstruction artifacts. Existing methods usually assume idealized, uniform single-axis rotation: they detect complete rotational cycles by periodic similarity matching of sequential projections and then assign angles by dividing the cycle into equal increments.~\cite{C7LC00943G, Pirone:21, Li:24}. In practice, however, non-spherical cells or cell aggregates exhibit complex rotational dynamics because asymmetric geometries experience non-uniform hydrodynamic shear. Besides, flow perturbations from syringe-pump pulsations, mechanical instabilities, microchannel imperfections, and partial fluid blockages also disrupt flow uniformity and rotational stability, even for spherical cells. As a result, a large fraction of cells with non-ideal motion patterns must be discarded, compromising sample representativeness and biasing population-level analysis. Moreover, in high-speed flows, cells often exit the imaging region before completing a full rotation, resulting in incomplete angular coverage. Additionally, cell-cell collisions, transient occlusions, and overtaking events also cause missing views. These issues transform the reconstruction into an ill-posed inverse problem with substantial spectral information loss, degrading reconstruction fidelity and narrowing the subset of cells compatible with conventional tomographic inversion.

Here we propose OmniFHT, a pose-free framework for 3D reconstruction in FHT under arbitrary unknown rotation trajectories. It unifies physics-based modeling and data-driven learning by iteratively refining the unknown pose (rotation and in-plane translation) and the volumetric RI distribution. For pose estimation, OmniFHT employs a coarse-to-fine strategy that progressively searches the pose and translation spaces (\(SO(3)\) and \(\mathbb{R}^2\)), conditioned on the current volumetric estimate to recover the optimal rotation and translation for each projection. For reconstruction, OmniFHT models the 3D scattering potential with an implicit neural representation (INR) in the Fourier domain, and optimize its network weights self-supervisedly by minimizing the data consistency loss between predicted and measured projections given the estimated poses~\cite{mescheder2019occupancy, mildenhall2021nerf, park2019deepsdf, sitzmann2020implicit}. The image formation model for data consistency is based on the Fourier diffraction theorem, which linearly transforms the 3D scattering potential to 2D projections via the Rytov weak-scattering approximation. By exploiting the implicit regularization embedded in compact neural representations, OmniFHT restores missing spectral information caused by limited angular coverage, collisions, and occlusions, enabling high-fidelity 3D reconstruction~\cite{yuce2022structured, zhong2021cryodrgn, zhong2021cryodrgn2, rangan2024cryodrgn, reed2021dynamic, shen2022nerp, song2023piner, molaei2023implicit}. 

We validated OmniFHT on simulated vacuolated cells and challenging experimental datasets, demonstrating substantial improvements in resolution and robustness. In simulations of vacuolated cells, OmniFHT achieves a 1.75× resolution improvement (measured by Fourier shell correlation, FSC~\cite{ludtke1999eman, liao2010definition}) over the standard Rytov-based reconstruction that assumes uniformly interpolated poses, while simultaneously recovering cells’ accurate rotational dynamics~\cite{Park2018,Chung2021,Ryu2021, kim2014high, kim2015simultaneous, lee2022inverse,Devaney:81,Sung:11,Lim:15,Gerchberg1974}. On experimental datasets of red blood cells (RBCs) and cell aggregates—systems that exhibit complex, multi-axis rotations—OmniFHT faithfully reconstructs 3D morphologies, whereas the standard Rytov-based method produces severe artifacts. It remains robust to realistic flow cytometry challenges such as collisions, occlusions, partial sliding, and incomplete angular coverage. Quantitative evaluation demonstrates that OmniFHT recovers accurate morphology from as few as five projections and resolves intracellular details with only 10 views, enabling high-throughput imaging under extreme sparsity. Even with one-third of the Fourier spectrum missing (120° angular coverage), it maintains high-fidelity 3D reconstructions. Furthermore, by eliminating the requirement to pre-select cells with single-axis rotation, OmniFHT, for the first time, reconstructed all cells in a clinical ascites specimen with diverse morphologies and rotational behaviors. This capability allows direct in situ characterization of cellular heterogeneity at the population scale, setting a new standard for unbiased, quantitative label-free cytometry.

\section{System and method}

\subsection{Optical system}

\begin{figure}[htbp]
    \centering
    \includegraphics[width=1.0\linewidth]{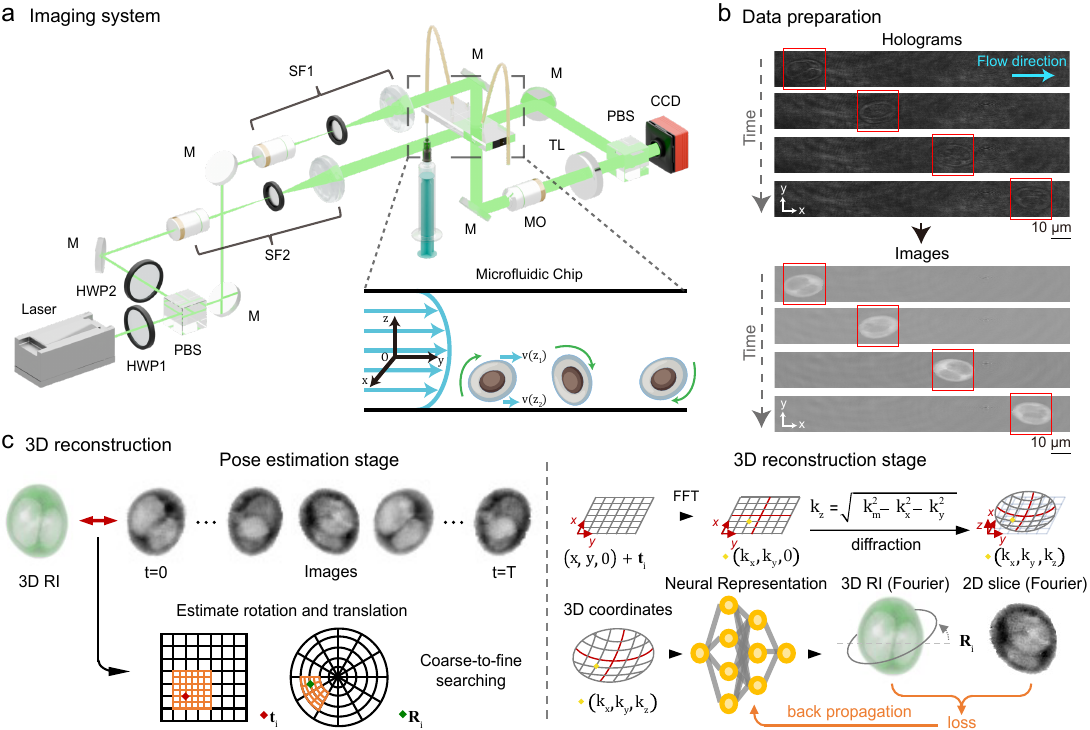}
    \caption{Overview of the OmniFHT pipeline: (a) Digital holographic flow cytometry acquires images under 532 nm plane-wave illumination as cells roll in a microfluidic chip; (b) Preprocessing converts raw hologram videos into projection sequences via refocusing, unwrapping, and denoising; (c) Iterative 3D reconstruction alternates between coarse-to-fine pose searching over $SO(3)$ and $\mathbb{R}^2$ grids to identify optimal rotation $\mathbf{R}_i$ and translation $\mathbf{t}_i$ for each image, and a physics-informed INR in the Fourier domain, grounded in the Fourier diffraction theorem, which is trained using self-supervised loss to recover high-fidelity 3D RI distributions.}
    \label{Fig. 1}
\end{figure}

We implement an FHT system based on a Mach-Zehnder off-axis interferometer to record holograms of flowing cells (Fig.\ref{Fig. 1}a). A microfluidic PMMA chip (Microfluidic-chip shop, Fluidic 144, Germany) enables high-speed recording of cell dynamics via single-exposure holography. A coherent laser (532 nm, 100 mW) serves as the light source and is split into object and reference beams by a polarizing beam splitter (PBS). Both beams are expanded and collimated into plane waves using spatial filters (SF) and a lens. A half-wave plate (HPW1) in front of the PBS is used to adjust the intensity ratio. Another half-wave plate (HPW2) employed in the reference arm makes the polarization direction of the reference beam consistent with the object beam for optimal interference. The object beam passes through the cell sample, positioned at the front focal plane of an infinity-corrected microscope objective (MO, 40×, NA = 0.7), and is imaged by the MO–tube lens (TL) pair to form a real image. Mirror M2 adjusts the reference beam path to generate off-axis interference with the object beam, forming a holographic pattern. This hologram is first formed through a beam splitter (BS) and then recorded by a CCD camera (5.86 $\mu$m pixel size, 1920×1200 resolution, PointGrey, Canada).

The cell suspension flows through the channel driven by a syringe pump, while the imaging system records holographic sequences of the flowing cells. The size of the microchannel, embedded within the chip, is 200$\mu$m × 200$\mu$m, the speed of the pump is set as 1$\mu$L/min, and the frame rate of the camera is 80 fps.

\subsection{Image formation model}

%The raw hologram \(I_H(x,y)\) is first used to recover the complex object field \(O(x,y)\), which encodes both amplitude and phase modifications imparted by the sample (Fig.~\ref{Fig. 1}b). Let \(O(x,y)\) and \(R(x,y)\) denote the object and reference waves, respectively. The recorded intensity is
Let \(O(x,y)\) and \(R(x,y)\) denote the object and reference waves, respectively. We illuminated the raw hologram \(I_H(x,y)=|O(x,y)+R(x,y)|^2\) with reference waves \(R\) to recover the complex object field \(O\), which encodes both amplitude and phase modifications imparted by the sample (Fig.~\ref{Fig. 1}b). The progress is

%\begin{equation}
%I_H(x,y)=|O(x,y)+R(x,y)|^2 = |O|^2 + |R|^2 + O R^* + O^* R.
%\end{equation}
\begin{equation}
R I_H = R (|O|^2 + |R|^2) + |R|^2 O + R^2 O^*\\
\end{equation}
In an off-axis holographic configuration, the cross-term \(|R|^2 O\) (the \(+1\) diffraction order) is spectrally separated from the zero order and the conjugate term in the Fourier domain. After applying a suitable bandpass filter in the Fourier plane to isolate this term, \iffalse compensating for the known reference \(R\),\fi and performing inverse propagation via the angular spectrum method, the full complex object field \(O\) is reconstructed up to a known scale factor. The phase map \(\phi_o(x,y)=\arg[O(x,y)]\) and amplitude map \(A_o(x,y)=|O(x,y)|\) can be extracted when necessary for visualization or downstream tracking.%, but the reconstruction pipeline and inverse modeling operate directly on the complex field.

Under the Rytov approximation, the complex perturbation of the field, defined as the logarithm of the normalized object field relative to the incident plane wave \(O_0\) (which we take as unit amplitude), i.e.,
\begin{equation}
\psi(x,y)=\ln\left[\frac{O(x,y)}{O_0(x,y)}\right],
\end{equation}
%is linearly related to the 3D scattering potential \(f(\mathbf{r})\). And the RI distribution \(n(\mathbf{r})\) is connected to \(f(\mathbf{r})\) by
%\begin{equation}\label{eq:scattering}
%n(\mathbf{r}) = n_m\sqrt{\frac{f(\mathbf{r})}{k_m^2}+1},
%\end{equation}
%where \(k_m=2\pi n_m/\lambda\) is the wavenumber in the background medium with RI \(n_m\) and \(\lambda\) the illumination wavelength. 

Given the complex field \(\psi(x,y)\) of the whole field of view (FOV), we derive the per-cell observation model. As cells traverse the FOV, automated tracking identifies regions of interest (ROIs) for individual cells. The measured complex field of a cell in each frame is affected by its spatial pose, which is parameterized by a 3D rotation \(\mathbf{R}\in\mathrm{SO}(3)\) and an in-plane 2D translation \(\mathbf{t}\in\mathbb{R}^2\). Denoting the 2D Fourier transform of \(\psi(x,y)\) by \(\hat{\psi}(K_x,K_y)\), the Fourier diffraction theorem establishes that the Fourier-domain scattering potential sampled under a given pose \((\mathbf{R},\mathbf{t})\) relates to \(\hat{\psi}\) on the Ewald sphere as:
\begin{equation}\label{eq:flow_image_formation}
\hat{f}\bigl(\mathbf{R}^{-1}\mathbf{k}\bigr)
=
2i\,k_z\,\hat{\psi}(K_x,K_y)\,
\exp\bigl(-i\,\mathbf{k}\cdot\mathbf{t}\bigr)
\exp\bigl[i\bigl(k_{0z}-k_z\bigr)z_0\bigr],
\end{equation}
with
\[
k_x = K_x + k_{0x},\quad
k_y = K_y + k_{0y},\quad
k_z = \sqrt{k_m^2 - k_x^2 - k_y^2},\quad
k_x^2 + k_y^2 \le k_m^2.
\]
Here, \(\hat{f}\) and \(\hat{\psi}\) denote the Fourier-domain representations of the 3D scattering potential and the complex Rytov perturbation, respectively. \(\mathbf{k}=(k_x,k_y,k_z)\) is the scattered wave vector, and \(\mathbf{k}_0=(k_{0x},k_{0y},k_{0z})\) is the incident illumination wave vector. In our imaging configuration the illumination is a plane wave propagating along the \(z\)-axis, so \(k_{0x}=k_{0y}=0\) and \(k_{0z}=k_m\). \(\hat{\psi}\) is linearly related to the \(\hat{f}\), and the RI distribution \(n(\mathbf{r})\) is connected to \(f(\mathbf{r})\) by
\begin{equation}\label{eq:scattering}
n(\mathbf{r}) = n_m\sqrt{\frac{f(\mathbf{r})}{k_m^2}+1},
\end{equation}
where \(k_m=2\pi n_m/\lambda\) is the wavenumber in the background medium with RI \(n_m\) and \(\lambda\) the illumination wavelength.

\subsection{Joint optimization of poses and volumetric reconstructions} 

While the image formation model outlined above enables reconstruction of the 3D RI distribution \(n\) from observed complex Rytov perturbations \(\{\psi_i\}_{i=1}^N\) with known poses, real flow cytometry experiments typically capture images as cells roll through microfluidic channels, resulting in unknown cellular poses. OmniFHT addresses this challenge through a unified framework that jointly recovers both the pose \(\omega_i = (\mathbf{R}_i, \mathbf{t}_i)\) for each observed perturbation \(\psi_i\) and the underlying 3D scattering potential \(f\) of the cell. Given no prior knowledge on either the 3D structure or the poses, we formulate the reconstruction as a maximum likelihood estimation (MLE) problem in Fourier space:
\begin{equation}
\{\hat f_{\mathrm{opt}}, \{\omega_i\}_{\mathrm{opt}}\}
= \arg\max_{\hat f,\{\omega_i\}} \sum_{i=1}^N \log p\bigl(\hat\psi_i \mid \hat f, \omega_i\bigr),
\label{eq:mle_formulation}
\end{equation}
where \(p(\hat\psi_i \mid \hat f, \omega_i)\) denotes the likelihood of observing the 2D complex Rytov perturbation \(\hat\psi_i\) given the 3D scattering potential \(\hat f\) in Fourier domain and pose \(\omega_i\). Within this probabilistic perspective, each measured \(\hat\psi_i\) furnishes constraints on the unknown 3D RI distribution while simultaneously allowing inference of the corresponding pose. To disentangle the coupled dependencies between RI reconstruction and cellular poses, we implement an iterative optimization framework alternating between volumetric reconstruction and pose estimation. 

Given a random initialization of cellular poses, we start from the volumetric reconstruction phase (Fig.\ref{Fig. 1}c, right). Rather than discretizing the cell’s RI on a fixed grid, we employ an INR \(f_\theta\) parameterized by \(\theta\) to represent the 3D scattering potential \(\hat f\) in the Fourier domain. Here, INR is a coordinate-based neural network that learns a mapping from frequency-space coordinates to the complex-valued scattering potential. Given the pre-defined image formation model, for each observed complex Rytov perturbation \(\hat\psi_{\mathrm{obs},i}\) with the current pose \(\omega_i\), we can derive the predicted perturbation \(\hat\psi_{\mathrm{pred},i}\) as:
\begin{equation}
\hat\psi_{\mathrm{pred},i}(K_x,K_y)
=
\frac{f_\theta\bigl(\mathbf R_i^{-1}\mathbf k\bigr)\,\exp\bigl(i\,\mathbf k \cdot \mathbf t_i\bigr)}{2\,i\,k_z}
\exp\bigl[-\,i\bigl(k_{0z} - k_z\bigr)\,z_0\bigr].
\label{eq:pred_ryotov_inr}
\end{equation}
Here, the network \(f_\theta\) takes rotated Fourier coordinates \(\mathbf R_i^{-1}\mathbf k\) as input and outputs the corresponding value of the scattering potential in the Fourier domain. Considering all observed perturbations, we train \(f_\theta\) in a self-supervised manner where the loss function is:
\begin{equation}
\mathcal{L}_\theta \;=\; \sum_{i=1}^N \bigl\lVert \hat\psi_{\mathrm{pred},i} \;-\; \hat\psi_{\mathrm{obs},i}\bigr\rVert_2^2, 
\quad i \in [1,N].
\label{eq:inr_loss}
\end{equation}

Once the scattering potential \(f\) has been updated, we proceed to refine the cellular pose for each observed perturbation \(\psi_{\mathrm{obs},i}\). OmniFHT employs a coarse-to-fine pose estimation strategy to efficiently search the joint space of 3D rotation \(\mathbf{R}\) and in-plane translation \(\mathbf{t}\). At the beginning of each pose estimation stage, a candidate pose set is constructed: a uniform grid over \(\mathrm{SO}(3)\) with \(30^\circ\) angular intervals and a translation grid over \([-0.5, 0.5]^2\) with spacing 0.1 (black grids in Fig.\ref{Fig. 1}c). For each candidate pose \((\mathbf{R}, \mathbf{t})\), we synthesize the corresponding predicted complex Rytov perturbation \(\psi_{\mathrm{pred},i}\) using the current estimate of \(f\) and the image formation model in Eq.~\eqref{eq:pred_ryotov_inr}. The complex cross-correlation between \(\psi_{\mathrm{pred},i}\) and the observed \(\psi_{\mathrm{obs},i}\) is computed as a similarity metric quantifying alignment fidelity. We retain the top eight pose hypotheses with the highest similarity scores. Around each of these, a finer local grid is constructed by bisecting both the rotational and translational resolutions (orange grids in Fig.\ref{Fig. 1}c), forming a new candidate pose set for the next refinement stage. This hierarchical search is repeated for five iterations, progressively narrowing the search space and improving pose accuracy. After the final iteration, the pose with the highest similarity score is selected as the estimated cellular pose for the corresponding measurement.

\subsection{Implementation details}

The INR used to represent the 3D scattering potential map is implemented as a multilayer perceptron (MLP) with three hidden layers, each containing 256 neurons. The input to the MLP is a 3D frequency-space coordinate \(\mathbf{k} = (k_x, k_y, k_z)\), normalized to lie within \([-0.5, 0.5]^3\). To capture both low- and high-frequency features of the scattering potential, we apply a sinusoidal positional embedding (PE) to each component of \(\mathbf{k}\) before passing it into the MLP. Specifically, for each scalar component \(k \in \{k_x, k_y, k_z\}\), we define a 32-dimensional embedding
\begin{equation}
\gamma(k) = \bigl[\sin(\alpha_1 k),\;\cos(\alpha_1 k),\;\sin(\alpha_2 k),\;\cos(\alpha_2 k),\;\dots,\;\sin(\alpha_{16} k),\;\cos(\alpha_{16} k)\bigr],
\label{}
\end{equation}
where \(\alpha_m = 2^{m-1}\pi\) for \(m = 1,2,\dots,16\). Thus each component \(k\) maps to 32 channels spanning multiple frequency scales. For the full 3D coordinate \(\mathbf{k}\), we concatenate these embeddings along each axis, yielding a 96-dimensional input:
\begin{equation}
\Gamma(\mathbf{k}) = \bigl[\gamma(k_x),\;\gamma(k_y),\;\gamma(k_z)\bigr].
\label{}
\end{equation}
The MLP then outputs a single scalar value corresponding to the scattering potential at the given frequency coordinate \(\mathbf{k}\).

Training process lasts 200 epochs with stochastic gradient descent (SGD) optimizer using a learning rate of 0.01 and a batch size of four. The model is implemented in PyTorch and trained on a single NVIDIA RTX 3090 GPU, with a typical memory footprint of approximately 17 GB. During training, the pose estimation is processed every five epoch, which is tested to be robust to jointly estimate accurate poses and reconstruct 3D scattering potential map.

\subsection{Cell culture and sample preparation}

In this study, we evaluated OmniFHT using two specimen types. The first specimen comprises the human urothelial bladder cancer cell line SW780, cultured in vitro to provide standardized conditions for method development. The second specimen is human ascites fluid obtained from patients, containing a heterogeneous mixture of cellular components—such as RBCs, WBCs, and ovarian cancer cells.

The SW780 cell line was procured from Beijing Biotides Biotechnology Co., Ltd. and authenticated prior to use. Cells were maintained in Leibovitz’s L-15 Medium (Gibco, \#11415064) supplemented with 10\% fetal bovine serum (FBS) (Gibco, \#10099141) and 1\% penicillin-streptomycin (Gibco, \#15140122) at 37℃ in 5\% CO$_2$ humidified incubator (Thermo HERAcell150i). After the treatment, cells were collected from wells using trypsin digestion, and dissolved in the dilution medium (PBS) with a cell density of 1×$10^5/mL$.

The ascites fluid was collected from ovarian cancer patients (200–500 mL), and the fresh sample was allowed to stand undisturbed at 4°C for 1 hour. Following sedimentation, we discarded the supernatant, washed the cell pellet with phosphate-buffered saline (PBS), and centrifuged it at 1,500g for 5 minutes, then remove the supernatant again. The resulting pellet was resuspended in Roswell Park Memorial Institute (RPMI) 1640 medium and adjusted to a final cell density of 1×$10^5/mL$ for subsequent flow cytometry analysis. Notably, all RBCs, WBCs, and ovarian cancer cells analyzed in this study were derived from the original ascites fluid.

\section{Results}

\subsection{Numerical Validations of OmniFHT}

\begin{figure}[htbp]
    \centering
    \includegraphics[width=1.0\linewidth]{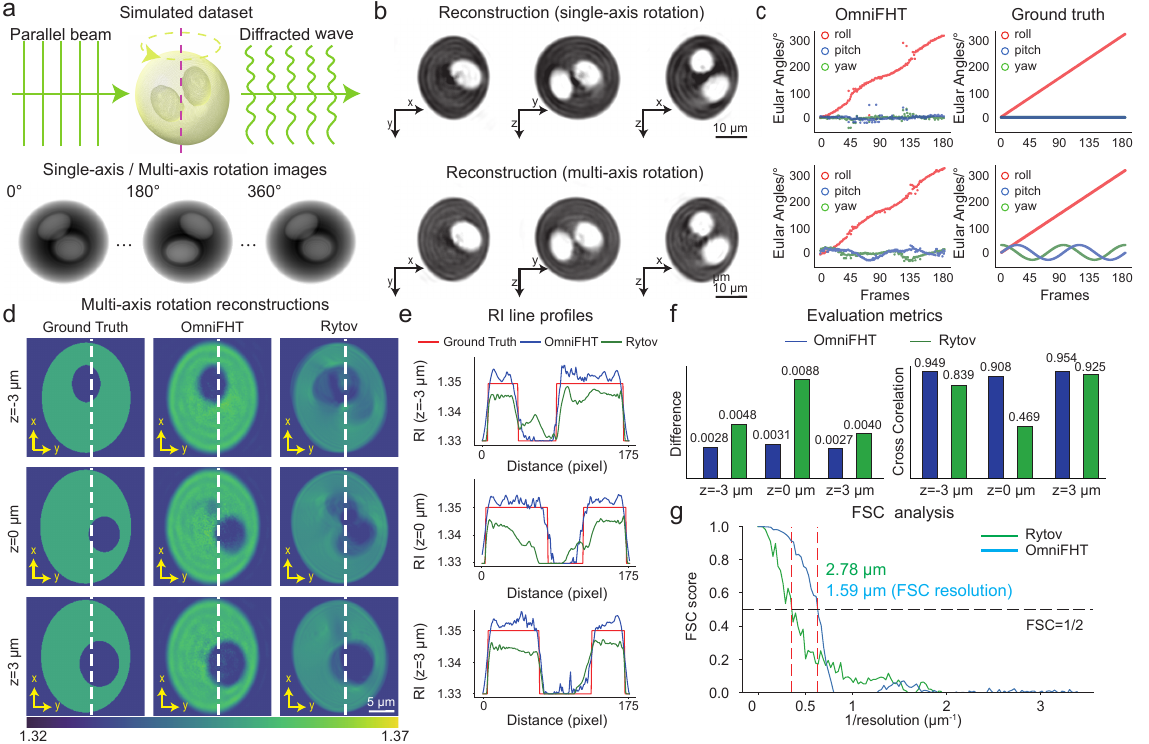}
    \caption{Simulated validation of OmniFHT: (a) BPM-generated datasets with a 532~nm plane-wave illumination on a simulated cell model; single-axis rotations use 180 projections at $2^\circ$ increments, multi-axis rotations use sinusoidal $30^\circ$ variations. (b) Orthogonal slices show that OmniFHT resolves internal structures under both rotation conditions. (c) Pose estimation accuracy is confirmed by close agreement between estimated and ground-truth Euler angles. (d) Under multi-axis rotations, OmniFHT outperforms a Rytov-approximation baseline, preserving vacuole boundaries at slices $z=-3~\mu m$, $z=0~\mu m$, and $z=3~\mu m$. (e–f) Line profiles and quantitative metrics demonstrate reduced errors and higher cross-correlation for OmniFHT. (g) FSC analysis indicates OmniFHT achieves 1.59 $\mu$m resolution versus 2.78 $\mu$m for the baseline, highlighting improved spatial fidelity.}
    \label{Fig. 2}
\end{figure}

We evaluated OmniFHT’s pose estimation and 3D RI reconstruction using a simulated cell (RI = 1.35) containing two internal vacuoles immersed in a medium (RI = 1.33), with diffraction patterns computed via the beam propagation method (BPM)\cite{kamilov2016optical}
%\textcolor{red}{(He: add BPM citations if you metion the term for the first time.)} 
under 532 nm plane wave illumination. The single-axis rotation dataset consists of 180 projections acquired by rotating the cell around the $z$-axis in 2° increments. We also generated a multi-axis rotation dataset where pitch and yaw angles vary sinusoidally with 30° amplitude and initial phases of 0 and $\pi/2$, respectively (Fig.\ref{Fig. 2}a). Figure~\ref{Fig. 2}b shows orthogonal cross-sections (at index 128 along each axis) of the reconstructed 3D RI distributions, where OmniFHT accurately recovers the internal vacuolar structures for both single- and multi-axis datasets. Pose estimation accuracy is validated by comparing estimated and ground truth rotation angles (Fig.\ref{Fig. 2}c). For single-axis rotation, the estimated roll angle increases linearly with frame number, while pitch and yaw remain near zero, confirming the expected motion pattern. For multi-axis rotation, OmniFHT successfully captures the sinusoidal oscillations in both pitch and yaw, demonstrating robust pose tracking under complex, non-uniform rotational dynamics.

To quantitatively evaluate reconstruction fidelity under multi-axis rotation, we benchmarked OmniFHT against the most commonly used Rytov-based method (see detailed implementation in Supplementary Information). Comparative analyses were performed on transverse slices at $z=-3~\mu m$, $z=0~\mu m$, and $z=3~\mu m$, where the slide $z=0~\mu m$ is located at the central slide of the reconstruction (Fig.\ref{Fig. 2}d).
%\textcolor{red}{(He: In all ODT/IDT/3D imaging papers, authors report z-distance using standard unit (e.g., $\mu$m), slice number is not helpful for understanding the experimental setting. Please correct this in all paragraphs and figures.)}. 
OmniFHT robustly recovers the vacuole boundary and spatial localization, with results closely matching the ground truth. In contrast, the baseline method introduces boundary distortions, including elongation and contraction around the vacuole, particularly near the margins. Notably, the Rytov-based method systematically underestimates RI values~\cite{Chung2021}, an artifact attributable to incomplete frequency coverage, whereas OmniFHT produces reconstructions whose overall morphology is consistent with the ground truth. 
%\textcolor{red}{(He: We talked about this last time. It is unnecessary to say that we over estimate the value. I would recommend that we directly claim the overall morphology matches GT.)} 
%Despite these biases, the overall morphology and spatial trends in the reconstructed RI distribution using OmniFHT remain in excellent agreement with the ground truth. 
Line profiles extracted from these layers revealed superior concordance between OmniFHT reconstructions and ground truth RI distributions, substantially surpassing the baseline method (Fig.\ref{Fig. 2}e). Quantitative analyses supported these qualitative comparisons, where at the slice corresponding to $z=0~\mu m$, OmniFHT reduced the average absolute difference by 64.77\% (from 0.0088 to 0.0031) and improved the cross-correlation coefficient by 93.60\% (from 0.469 to 0.908) relative to the baseline (Fig.\ref{Fig. 2}f). Comparable improvements were also observed at all shown layers.
%, with reductions of 41.67\% and 13.11\% at z=105, and 32.50\% and 3.14\% at z=145, for the two metrics, respectively. 
%\textcolor{red}{(He: I didn't understand the meaning of two numbers in "with the reductions of 41.67\% and 13.11\%" and "32.50\% and 3.14\%'".)} 
Furthermore, Fourier Shell Correlation (FSC) analyses (Fig.\ref{Fig. 2}g, see SI for details) demonstrated that OmniFHT achieved a spatial resolution of 1.59~$\mu$m, indicating enhanced reconstruction detail. In comparison, the baseline method achieved a spatial resolution of 2.78~$\mu$m, corresponding to an improvement factor of approximately 1.75-fold with OmniFHT. Notably, in simulations we report correlation against the ground truth using the 1/2 criterion; in experiments we report half-set FSC with the 1/7 criterion. This choice follows common practice for reference-based vs split-data evaluations, respectively\cite{van2005fourier}.
%\textcolor{red}{(He: citations.)}

\subsection{Experimental validations of OmniFHT}

\subsubsection{OmniFHT adapts to realistic single-axis rotation}

In real-world biological samples, complex multi-axis rotational dynamics are often coupled with morphologically intricate cells, as their structural complexity leads to non-uniform force distributions within microfluidic flow fields. We applied OmniFHT to experimentally acquired datasets, including both cells undergoing near-uniaxial rotation and those with complex multi-axis motion.

\begin{figure}[htbp]
    \centering
    \includegraphics[width=1.0\linewidth]{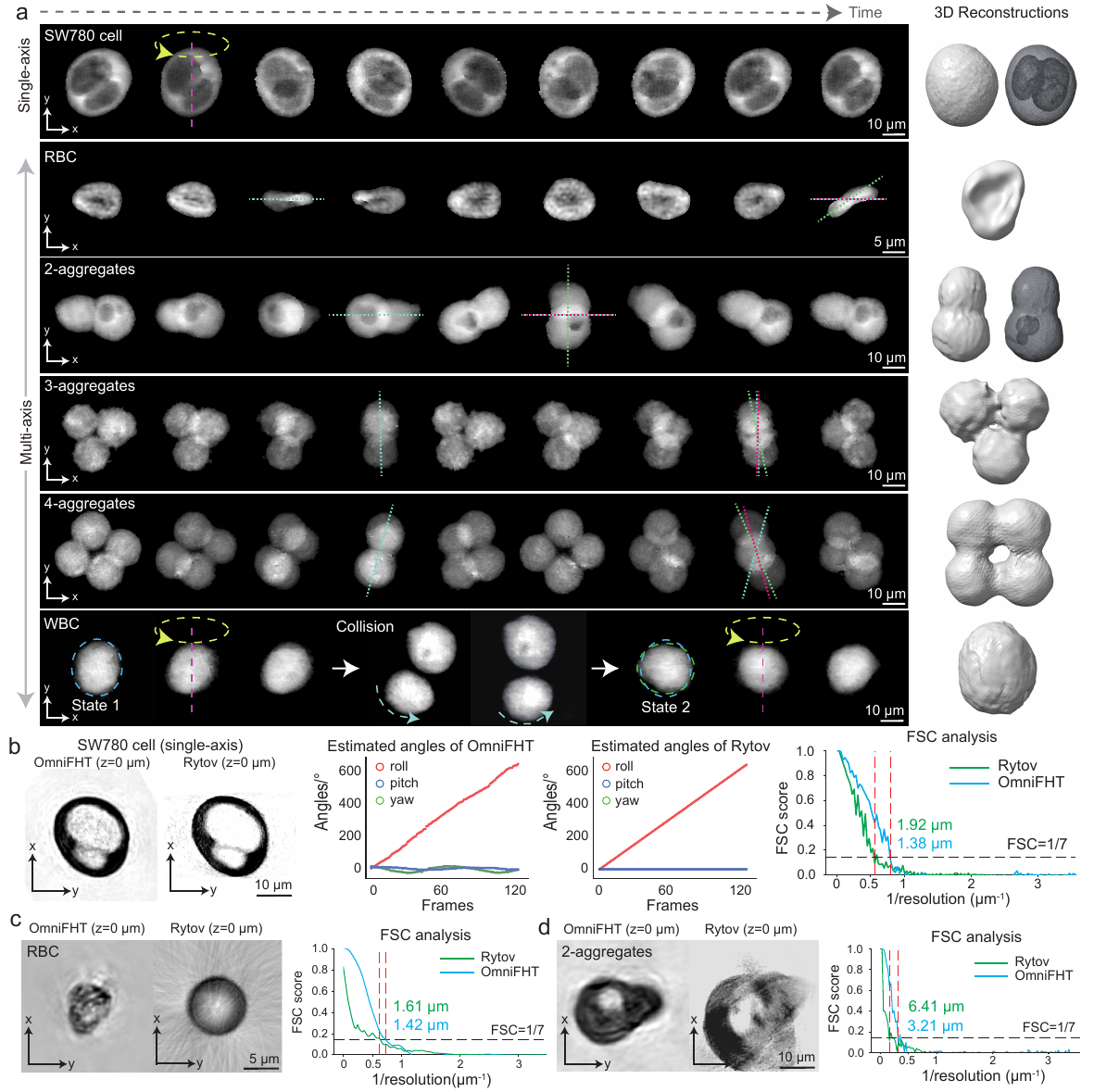}
    \caption{OmniFHT enables high-fidelity 3D RI reconstruction of cells undergoing complex, multi-axis rotations. (a) Six representative time-series sequences of projections from one SW780 cell, one RBC, three SW780 aggregates (2-, 3-, and 4-cell), and one WBC. Each sequence shows consecutive projections (left) and corresponding 3D RI reconstructions (right). Yellow arrows (top left) indicate shear-driven rotation. Blue and green dashed lines mark the principal axes at 0° and 180° rotation, and the red line is the in-plane symmetry axis of the blue. Misalignment between green and red reveals out-of-plane rotational components, violating single-axis assumptions. For the WBC in the last row, a collision with another WBC induced relative rotation, resulting in a marked change of the rotational axis, as indicated by the blue and green dashed ellipses. (b) For a vacuolated SW780 cell undergoing near-uniaxial rotation, OmniFHT recovers morphology and pose estimates comparable to Rytov-based reconstruction, and achieves better FSC resolution. (c) For an RBC with pronounced multi-axis rotation, only OmniFHT resolves the characteristic biconcave morphology in the x–y plane. FSC analysis confirms superior resolution (1.42~$\mu$m) over the baseline (1.61~$\mu$m). (d) For the two-cell aggregate case, the Rytov-based method fails due to incorrect pose estimation, while OmniFHT improves spatial resolution from 6.41~$\mu$m to 3.21~$\mu$m.}
    \label{Fig. 3}
\end{figure}

We first examined a spherical vacuolated SW780 cell (123 frames, 256×256 pixel size) exhibiting shear-driven, predominantly uniaxial rotation around the $z$-axis, as indicated by the aligned principal axes across the time series (Fig.~\ref{Fig. 3}a, top row). The dominant rotation direction is marked by the red axis and yellow arrows in the projection panel. OmniFHT accurately estimated the rotational trajectory and reconstructed the 3D RI distribution with high fidelity, clearly resolving the two internal vacuoles—demonstrating its robustness even under idealized motion assumptions (Fig.~\ref{Fig. 3}b). 

For quantitative evaluation, we split the 123-frame SW780 cell dataset into two independent half-sets (61- and 62-frames), performed separate 3D reconstructions, and computed the FSC between them using the standard FSC = 1/7 threshold as the resolution criterion (see SI for details). OmniFHT achieved an FSC resolution of 1.38~$\mu$m, superior to the 1.92~$\mu$m resolution of the Rytov method, demonstrating significantly enhanced capability for single-axis rotational samples.

\subsubsection{OmniFHT under Complex Multi-Axis Motion}

Next, we tested OmniFHT on real complex datasets of multicellular aggregates and irregularly shaped RBCs, including two-cell (116 frames), three-cell (112 frames), and four-cell (57 frames) SW780 cell aggregates, each captured at 256×256 pixel size, as well as one red blood cell (RBC) dataset with 106 projection images (64×64 pixels). These datasets represent diverse biologically relevant configurations, ranging from single cell to tightly packed multicellular aggregates with overlapping scattering signatures, both of which pose significant challenges for conventional tomographic reconstruction approaches. Figure~\ref{Fig. 3}a (rows 2–5) presents four representative time-series sequences, each showing nine selected projection images (left) and their corresponding 3D RI reconstructions by OmniFHT (right). To assess rotational behavior, we analyzed the alignment of principal cell axes across the sequence. In each case, the blue and red dashed lines mark the principal axis at the beginning of the rotation and its in-plane symmetric counterpart after a 180° uniaxial rotation, respectively—i.e., the expected orientation under ideal single-axis motion. The green dashed line indicates the actual principal axis at the 180° time point. A clear misalignment between the green (observed) and red (expected) axes reveals significant out-of-plane rotational components, demonstrating that all four samples undergo complex, multi-axis rotations that violate the assumptions of conventional FHT methods.

For the RBC dataset, OmniFHT successfully reconstructed a high-fidelity 3D RI distribution that exhibited a slightly asymmetric, triangularly skewed biconcave elliptical morphology, in close agreement with the observed cell appearance in both the projections and the final 3D reconstruction. And as shown in Fig.~\ref{Fig. 3}c, The cross-sectional view at the central z-slice of the 3D RI reconstruction produced by OmniFHT also reveals a morphology consistent with the observed cell appearance, namely a triangularly skewed biconcave ellipse. In contrast, the baseline Rytov-based reconstruction yielded a nearly spherical shape with central hollowing and artifacts, which is highly inconsistent with the expected cellular morphology. For quantitative assessment of reconstruction fidelity with the half-set FSC analysis (1/7 criterion), OmniFHT achieved a spatial resolution of 1.42~\(\mu\)m, outperforming the baseline method’s resolution of 1.61~\(\mu\)m. Notably, in the low-frequency region of the FSC curve, the baseline method exhibits markedly lower similarity, which arises from artifacts induced by angular estimation errors. By contrast, OmniFHT effectively mitigates these artifacts and preserves consistency in the low-frequency regime.

Beyond single-cell imaging, OmniFHT’s ability to resolve complex multi-cellular configurations was further evaluated using datasets of SW780 cell aggregates. In Fig.~\ref{Fig. 3}a (rows 3–5), OmniFHT successfully reconstructed the 3D RI distributions with high fidelity for each case, accurately resolving individual cell boundaries and internal 3D RI distributions in the presence of overlapping scattering signals and multi-axis rotational dynamics. Figure~\ref{Fig. 3}d further presents representative cross-sectional views of the two-cell aggregate by OmniFHT and Rytov-based method. In contrast to the Rytov-based baseline method which produced blurred boundaries and artifacts due to incorrect rotation estimation, OmniFHT successfully resolved individual cell boundaries and internal vacuole structures with high fidelity. Moreover, surface-mesh visualizations of the 3D RI distributions further validated OmniFHT's precision in reconstructing internal structures, most notably the two adjacent spherical vacuoles, which highlight its robustness in resolving fine-scale cellular features. To quantitatively assess the spatial resolution of this reconstruction, OmniFHT improves the half-set FSC resolution (1/7 criterion) to 3.21~\(\mu\)m, significantly improving upon the baseline method’s resolution of 6.41~\(\mu\)m, with a 2.00-fold enhancement. This improvement highlights OmniFHT’s superior capability in preserving structural detail in multi-cellular systems where traditional approaches fail due to model mismatch and rotational ambiguity. 

\subsubsection{OmniFHT for multi-axis rotations induced by cell–cell interactions} 

Beyond shear-induced multi-axis rotations arising from intrinsic cell morphology or fluid perturbations, high-throughput flow cytometry also encounters frequent cell–cell collisions, which represent a primary source of rotational complexity. For the white blood cell (WBC, last row in Fig.~\ref{Fig. 3}a), during the collision, the interacting cells underwent relative rotation, as indicated by the dashed arrows. This rotation induced a shift in the in-plane rotational axis—from vertical to horizontal (blue and green elliptical dashed lines)—thereby generating multi-axis dynamics. From 65 projections, OmniFHT reconstructed an ellipsoidal 3D RI distribution consistent with the observed morphology. This highlights OmniFHT’s robustness in handling collision-induced rotational heterogeneity, a common challenge in realistic flow cytometry conditions.

\subsubsection{OmniFHT for sparse view and limited angle reconstruction}
%\textcolor{red}{(He: 3.2.4 and 3.2.5 belong to the same topic. Please merge into one single subsection.)}
In practical rapid, high-throughput FHT scenarios, achieving accurate 3D RI reconstructions often requires acquiring a relatively large number of projection images per cell, typically 100–400 frames. However, such high acquisition rates can be challenging to achieve due to constraints imposed by hardware limitations, high flow velocities, and occlusion or collision between cells. 

\begin{figure}[htbp]
    \centering
    \includegraphics[width=1.0\linewidth]{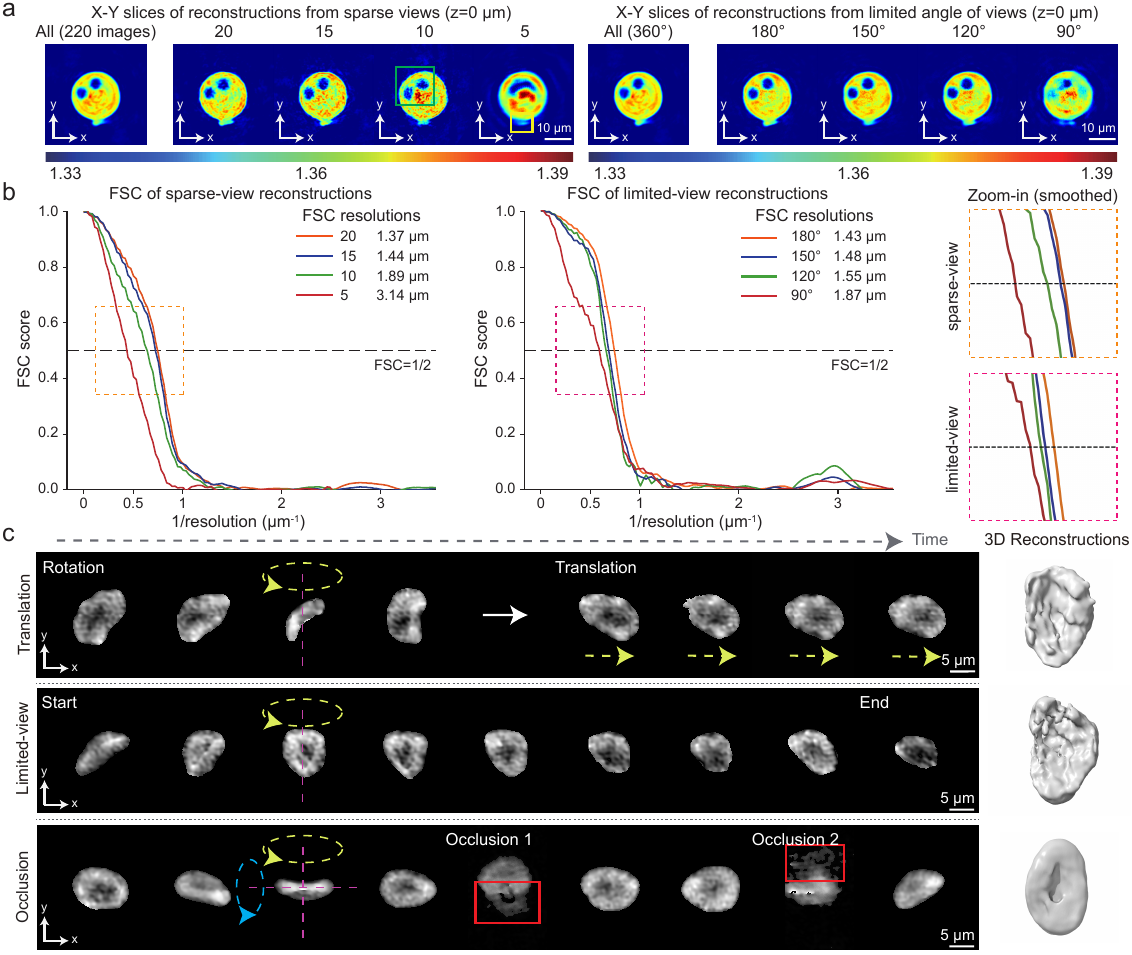}
    \caption{
    (a) Left: sparse-view reconstructions of a SW780 cell in the x–y plane using subsets of projections uniformly sampled from the full 360° dataset containing 220 projections, each with 20, 15, 10, and 5 views. With only 10 views, key structures remain distinguishable (green boxes); at 5 views, intracellular details degrade but overall morphology and protrusions are still recognizable (yellow boxes). Right: Limited-angle reconstructions using angular spans of decreasing coverage: 180° (110 views), 150° (92 views), 120° (73 views), and 90° (55 views). While reconstructions at 180°, 150°, and 120° preserve internal vacuoles and boundary morphology, the 90° reconstruction shows visible degradation and blurring, consistent with the increasing severity of the missing wedge problem. 
    (b) FSC analyses for sparse-view (left) and limited-view (middle) reconstructions against the full 220-view reference. Zoomed-in views of central frequency regions highlight the relative performance differences between sparse- and limited-view settings (right). 
    (c) Real-world sparse-view or limited-view scenarios and reconstructions in high-throughput cytometry, including translation-dominated motion, flow dynamics limiting the angular span, and cell occlusion. 
    }
    \label{Fig. 4}
\end{figure}

First, We investigated OmniFHT’s effectiveness under sparse-view conditions by re-examining a vacuolated SW780 bladder cancer cell dataset (comprising 220 projections, covering a full 360° rotation, 256×256 pixels). To emulate practical scenarios involving fast and high-throughput imaging protocols, we uniformly subsampled the original dataset into progressively sparser subsets containing only 20, 15, 10, and 5 projections, respectively. As the number of views decreased incrementally from 20 down to 5, reconstruction quality gradually diminished (Fig.\ref{Fig. 4}a, left). Nevertheless, at the notably sparse condition of 10 projections, OmniFHT successfully reconstructed key cellular features, including two prominent intracellular vacuoles in the green bounding box in Fig.\ref{Fig. 4}a (left), in excellent correspondence with the full-view reference reconstruction. Even with only five input views, although the vacuole structures became blurred, the peripheral cell protrusion indicated by the yellow bounding box remained discernible, and the overall cellular boundary was largely preserved. For quantitative evaluation of reconstruction fidelity under these sparse-view conditions, we computed FSC curves between the sparse-view reconstructions and the reference reconstruction obtained from the full dataset, adopting an FSC threshold of 0.5 as an indicator of approximate spatial resolution. FSC-based resolutions were determined as 1.37~$\mu$m, 1.44~$\mu$m, 1.89~$\mu$m, and 3.14~$\mu$m for the 20-, 15-, 10-, and 5-view reconstructions, respectively (Fig.\ref{Fig. 4}b, left). These results highlight OmniFHT’s robustness in maintaining high-quality 3D reconstructions even under significantly sparse observational input. Although the experiments under extreme conditions led to oscillations in the FSC curves, the zoomed-in profiles clearly preserve the overall trend (Fig.\ref{Fig. 4}b, right).

Beyond sparse-view challenges, FHT systems also face constraints that limit angular coverage. Such limitations can arise from translational motion across the field of view or mutual occlusion between densely packed cells, leading to incomplete acquisition of projection angles. Unlike sparse-view sampling, which reduces the number of projections while preserving broad angular distribution, angularly constrained imaging truncates the accessible frequency domain and introduces pronounced missing wedge artifacts in Fourier space. We tested reconstructions of the vacuolated SW780 cell using subsets of the full 220-view dataset restricted to angular spans of 180° (using first 110 images), 150° (92 images), 120° (73 images), and 90° (55 images). These correspond to none, 1/6, 1/3, and 1/2 missing frequency content (w.r.t the full-view dataset), respectively. As shown in Fig.~\ref{Fig. 4}a (right), central slices of reconstructions at 180°, 150°, and 120° retained key morphological features—including two internal vacuoles and clear boundaries, whereas the 90° result exhibited notable degradation, with blurred intracellular details and suppressed contrast. FSC analysis against the full 220-view reference (Fig.~\ref{Fig. 4}b, middle) further quantified this trend, yielding spatial resolutions of 1.43~\(\mu\)m (180°), 1.48~\(\mu\)m (150°), 1.55~\(\mu\)m (120°), and 1.87~\(\mu\)m (90°) at the 0.5 threshold. Despite increasing spectral loss, OmniFHT effectively compensated for moderate angular truncation via its inherent spatial regularization, producing high-fidelity reconstructions even with up to one-third missing spectral content. 

We further applied OmniFHT to realistic scenarios with diverse imaging complications, containing three RBCs (two echinocytes and one discocyte) samples. Fig.~\ref{Fig. 4}c depicts three prevalent issues encountered in high-speed tomographic imaging: motion dominated by translation, limited angular acquisition range, and mutual occlusion. For the echinocyte in the top row, translational motion dominated over rotation, resulting in insufficient rotational coverage. Despite limited sampling, OmniFHT reconstructed the 3D RI distribution from 106 projections and resolved surface protrusions and concave features matching the observed morphology. Another echinocyte, imaged at high velocity, yielded only 22 projections with less than 180° of rotation due to insufficient dwell time in the field of view. As shown in the second row, OmniFHT preserved key morphological features, including the spiculated surface structure. In the final case (bottom row), mutual occlusion between densely packed cells caused partial blockage of the optical path, marked by red bounding boxes. After excluding occluded frames, the remaining 97 projections of the RBC were used for reconstruction. OmniFHT preserved the discocyte’s biconcave morphology, demonstrating resilience to partial obscuration. By systematically addressing these real-world challenges, we validated OmniFHT’s capability to produce high-fidelity 3D RI reconstructions under highly constrained conditions. These results highlight OmniFHT’s potential as a versatile tool for rapid, high-throughput cytometry applications, where traditional methods often fail due to incomplete angular coverage or complex cell interactions. 

\subsection{In situ volumetric cytometry for cellular populations in real clinical biofluids}

\begin{figure}[htbp]
    \centering
    \includegraphics[width=1.0\linewidth]{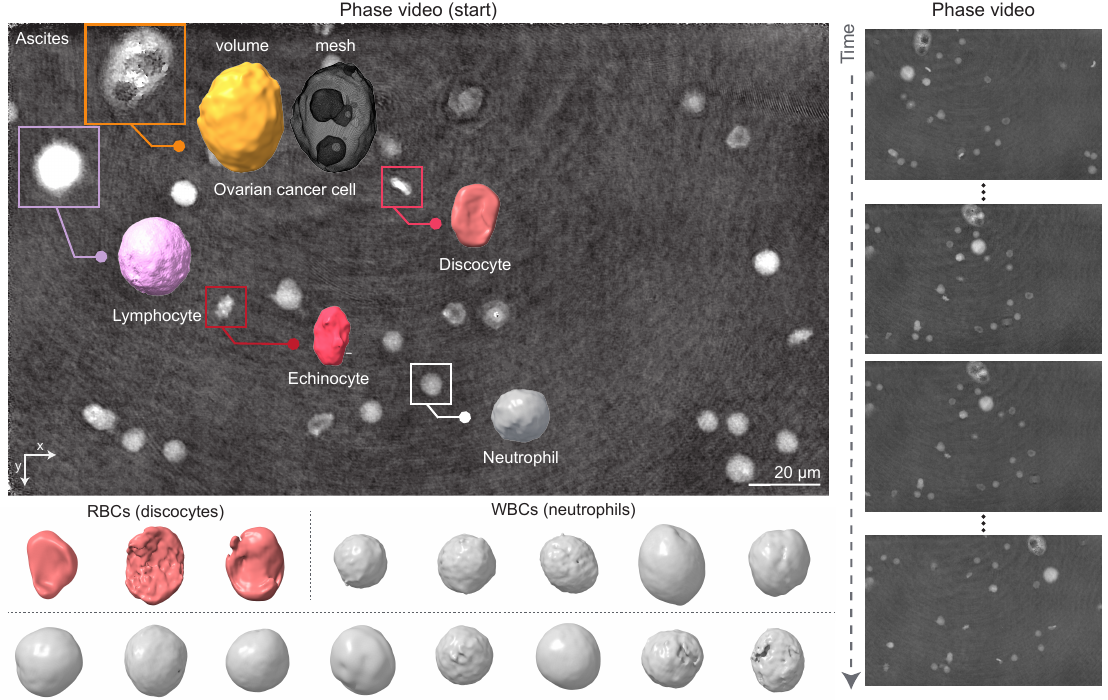}
    \caption{The figure demonstrates the 3D RI reconstructions of heterogeneous cellular populations in real clinical ascites using OmniFHT. The left upper corner shows the first phase video frame, highlighting representative examples and corresponding reconstructions of each cell type within the field of view: an ovarian cancer cell (orange bounding box), a lymphocyte (purple), a neutrophil (white), and two red blood cells (RBCs)—a discocyte and an echinocyte (red). In total, the video captured 21 cells within the focal plane. Below, the 3D RI reconstructions of the remaining cells in the population are shown, including three additional discocytes (red) and 13 neutrophils (white). On the right, a time-series phase video visualizes the complete sequence, revealing how fluid flow induces passive motion and complex rotational dynamics in the cells.}
    \label{Fig. 5}
\end{figure}

To extend the clinical applicability of OmniFHT in population-level analysis for unprocessed biofluids, we applied our method to a real clinical ascites sample under standard digital holographic flow cytometry conditions using 532 nm plane wave illumination. We captured a video containing a heterogeneous population of cells from clinical ascites specimens, containing RBCs, WBCs, and epithelial ovarian cancer cells. After excluding out-of-focus cells, as well as cells that were already located near the downstream end of the microfluidic chip at the first frame of acquisition (therefore contributing only a limited number of frames), the dataset retained 21 cells in total: one ovarian cancer cell, five RBCs (one echinocyte and four discocytes; 64×64), and 15 WBCs (14 neutrophils and one lymphocyte; mostly 64×64). Larger cells (cancer cell, lymphocyte) were captured at 256×256.

As shown in Fig.~\ref{Fig. 5}, a representative segment from the recorded phase video captures the holographic projections of these cells flowing through the microfluidic channel. In the initial frame (Phase video – start), four representative cells are highlighted: an epithelial ovarian cancer cell (orange bounding box), a discocyte and an echinocyte (red bounding boxes), a neutrophil (white bounding box), and a lymphocyte (purple bounding box). These cells experience natural, passive rotations in 3D rotation space, underscoring the intrinsic diversity in both cellular morphology and rotation trajectory across the population. Without requiring any rotational alignment constraints or trajectory priors, OmniFHT robustly recovered the 3D RI distributions and distinct morphological features for each individual cell. Notably, the epithelial ovarian cancer cell—outlined in orange exhibits an ellipsoidal shape with pronounced surface undulations and internal RI heterogeneity. Its black mesh rendering reveals two distinct internal cavities, likely reflecting vacuolization or nucleus–cytoplasm segregation—features consistent with the phase delays observed in the corresponding projections, thereby confirming the biological validity of the reconstruction. Among the RBCs, the echinocyte displays a spiculated contour, in contrast to the four discocytes which retain their canonical near-biconcave morphology with symmetric, gently curved surfaces. For WBCs, all 14 neutrophils and the lymphocyte exhibit smooth, approximately spherical shapes, with individual differences in surface roughness and size preserved in the reconstructions. 

In contrast to traditional approaches that often discard cells with non-single-axial rotational trajectories, OmniFHT imposes no trajectory assumptions or filtering. This key feature enables unbiased morphological reconstruction and statistical analysis of the entire cellular population, thereby providing a robust framework for downstream applications in phenotype stratification, rare cell detection, and high-throughput, label-free cytometry.

\section{Discussion}

In this study, we introduce OmniFHT, a unified pose-free approach for 3D cell and cellular population QPI under flow cytometry conditions. It directly addresses a longstanding challenge in FHT: reconstructing 3D RI distributions from projections acquired at unknown poses with complex rotation trajectories. Validated on both simulated and real datasets, OmniFHT demonstrates robust performance even with sparse/limited view sampling. By eliminating the constraint of single-axis rotation assumed in traditional methods, it substantially broadens the flexibility and applicability of FHT for 3D RI reconstruction, thereby making the results of  downstream applications such as cell detection and classification more reliable.

One key innovation of OmniFHT lies in its use of a coordinate-based INR to model the scattering potential field. This compact, grid-free representation, implemented as a multilayer perceptron (MLP) with only three hidden layers of 256 neurons each, inherently enforces smoothness and continuity as a form of regularization. This allows OmniFHT to compensate for missing frequency components under limited angular coverage or extreme sparsity, effectively recovering structural details even in the presence of severe spectral incompleteness. Furthermore, by using the Fourier diffraction theorem to fill the frequency spectrum, we enforce physical consistency and avoiding full 3D gradient backpropagation, substantially accelerate training.

Most significantly, OmniFHT enables, for the first time, unbiased volumetric reconstruction of all cells within continuously flowing clinical biofluids. Unlike traditional approaches that discard a substantial fraction of cells with complex rotation trajectories, our method reconstructs every encountered cell in situ, regardless of shape, pose trajectory, or biofluids' environments. This capability is particularly critical in diagnostic and screening contexts. By eliminating the need for preselection or filtering, OmniFHT mitigates sampling bias and recovers the full morphological spectrum of the cellular population. This establishes a foundation for population-scale, label-free RI cytometry, enabling systematic morphological analysis across heterogeneous cell populations and pointing toward future applications in early diagnostics and clinical cytometry.

Nonetheless, several limitations warrant future attention. Although our model exhibits strong performance across diverse scenarios, its pose estimator may lack robustness under extremely limited-angle or other challenging imaging conditions. Incorporating assumptions on rotational dynamics or cellular morphological priors could further enhance the precision and robustness of orientation estimation. Besides, cells located near the edges of the microfluidic channel exhibit degraded phase recovery due to optical distortions and reduced signal-to-noise ratios, and currently lack reliable strategies for accurate 3D reconstruction. From a deployment perspective, integration with microfluidic hardware for real-time inference will be essential to fully unlock its potential in clinical and pharmaceutical environments.\\

\noindent\textbf{Funding Statement.} This study was partially supported by National Key Research and Development Program of China (2022YFC3401100). \\

\noindent\textbf{Disclosures.} The authors have no conflicts of interest. \\

\noindent\textbf{Data availability.} Data will be made available on request.

\bibliography{sample}

\end{document}